\documentclass{article}
\DeclareUnicodeCharacter{FF1A}{:}

\usepackage[preprint]{neurips_2025}

\usepackage{hyperref}       
\usepackage{url}            
\usepackage{booktabs}       
\usepackage{amsfonts}       
\usepackage{amsmath}        
\usepackage{nicefrac}       
\usepackage{algorithm}
\usepackage{algorithmic}
\usepackage{microtype}      
\usepackage{xcolor}         
\usepackage[pdftex]{graphicx}
\usepackage{subcaption}
\usepackage{wrapfig}
\usepackage{multirow}
\usepackage{graphicx}
\usepackage{amsmath}
\usepackage{amsfonts}
\usepackage{amssymb}
\usepackage{booktabs}
\usepackage{xcolor}

\title{A Wavelet-based Stereo Matching Framework for Solving Frequency Convergence Inconsistency }

\author{
Xiaobao Wei$^{1,2}$, Jiawei Liu$^{1,3,4*}$, Dongbo Yang$^{1,3,4,7}$, Junda Cheng$^{5}$, Changyong Shu$^{6}$, Wei Wang$^{1,3,4}$\\
$^1$Shenyang Institute of Automation, Chinese Academy of Sciences\\ 
$^2$Nanjing University of Science and Technology, $^3$Liaoning Liaohe Laboratory\\ 
$^4$Key Laboratory on Intelligent Detection and Equipment Technology of Liaoning Province\\
$^5$Huazhong University of Science and Technology, $^6$Beihang University\\
$^7$University of Chinese Academy of Sciences\\
\texttt{wxb@njust.edu.cn, liujiawei@sia.cn}
}

\begin{document}

\maketitle

\begin{abstract}
We find that the EPE evaluation metrics of RAFT-stereo converge inconsistently in the low and high frequency regions, resulting high frequency degradation (e.g., edges and thin objects) during the iterative process. The underlying reason for the limited performance of current iterative methods is that it optimizes all frequency components together without distinguishing between high and low frequencies. We propose a wavelet-based stereo matching framework (Wavelet-Stereo) for solving frequency convergence inconsistency. Specifically, we first explicitly decompose an image into high and low frequency components using discrete wavelet transform. Then, the high-frequency and low-frequency components are fed into two different multi-scale frequency feature extractors. Finally, we propose a novel LSTM-based high-frequency preservation update operator containing an iterative frequency adapter to provide adaptive refined high-frequency features at different iteration steps by fine-tuning the initial high-frequency features. 
By processing high and low frequency components separately, our framework can simultaneously refine high-frequency information in edges and low-frequency information in smooth regions, which is especially suitable for challenging scenes with fine details and textures in the distance. Extensive experiments demonstrate that our Wavelet-Stereo outperforms the state-of-the-art methods and ranks $1^{st}$ on both the KITTI 2015 and KITTI 2012 leaderboards for almost all metrics. We will provide code and pre-trained models to encourage further exploration, application, and development of our innovative framework (\href{https://github.com/SIA-IDE/Wavelet-Stereo}{https://github.com/SIA-IDE/Wavelet-Stereo}).

\end{abstract}

\begin{figure}[htbp]
    \centering
    \includegraphics[width=\linewidth]{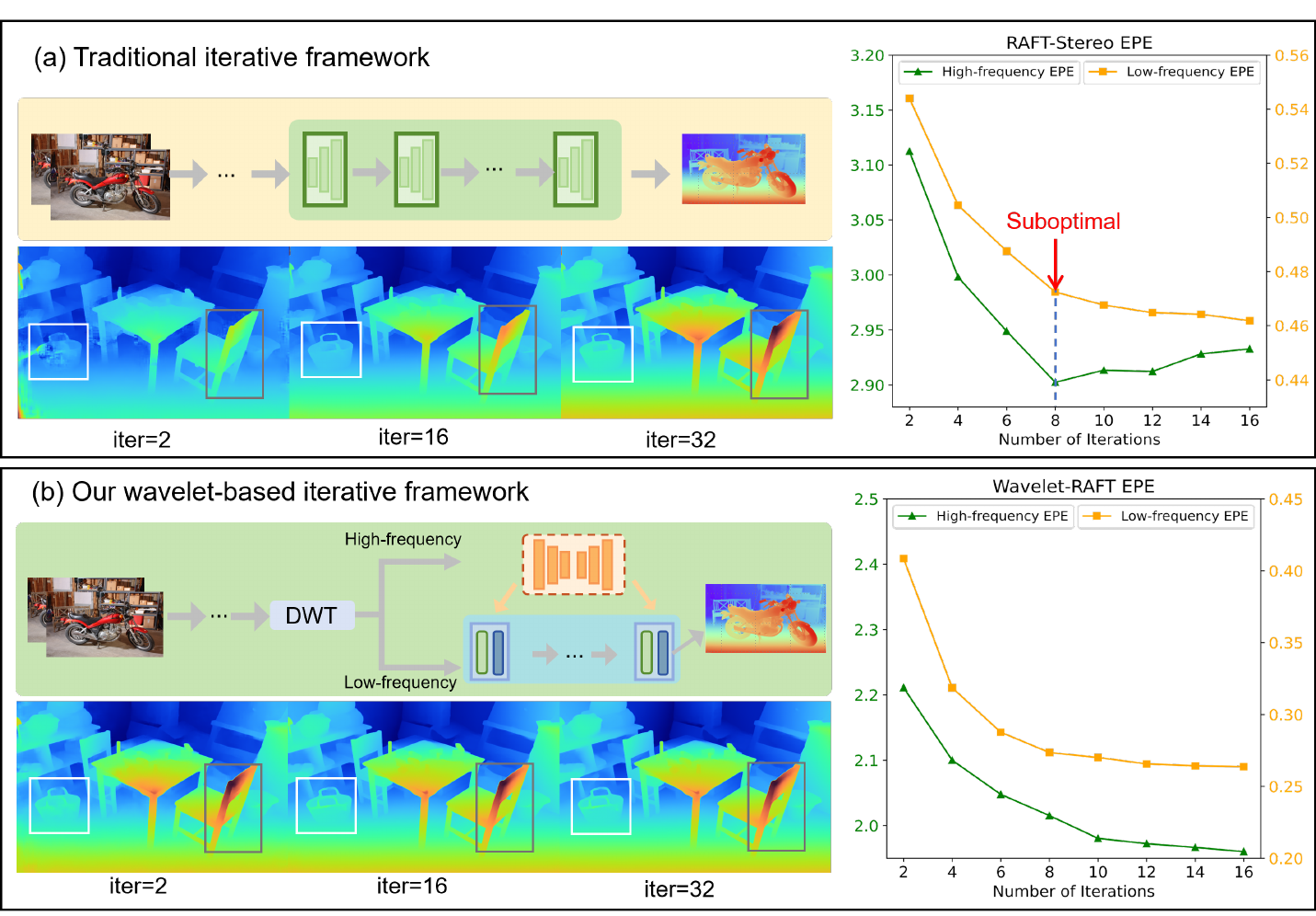}
    \caption{High and low frequency region EPE performance evaluation for some challenging scenes on ETH3D dataset~\cite{30}. (a) Traditional iterative-based methods~\cite{52} process the all frequency components uniformly, resulting in inconsistent convergence in different frequency regions. (b) We design frequency-specific feature extraction and processing modules to achieve overall optimization for different frequency components.}
  
    \label{fig:fengmian}
\end{figure}

\section{Introduction}
\renewcommand{\thefootnote}{\fnsymbol{footnote}}
\footnotetext[1]{Corresponding author.}
\renewcommand{\thefootnote}{\arabic{footnote}}

\label{section:1}
Stereo matching aims to estimate dense disparity maps by matching corresponding pixels between rectified stereo images. This technique serves as the cornerstone for autonomous driving~\cite{82}, augmented reality~\cite{80}, and robotic manipulation~\cite{81}. Despite decades of research, achieving high-precision and high-efficiency stereo matching remains challenging.

The advent of deep learning has revolutionized the field enabling end-to-end disparity prediction through CNNs~\cite{cheng2024adaptive, 84, 35, 85, 86, 87,88}. 
Aggregation-based methods~\cite{32, 36, 44, 40} improve accuracy by building 4D correlation volumes and applying 3D convolutions for regularization.
To avoid expensive 3D convolution, RAFT-stereo~\cite{52} updates the disparity map and hidden states by iteratively indexing from the all-pairs correlation volume and using the gate recursive unit operator.
Subsequent RAFT-based iterative methods~\cite{70, 55, 52, 50, 54, 73} suffer from the issue that correlation volume contains significant redundant noise, resulting high frequency degradation (e.g., edges and thin objects~\cite{50,54,73}) during the iterative process.
To address this, Selective-Stereo~\cite{50} employs variable receptive fields instead of a fixed one in the GRU to better capture high-frequency information, while DLNR~\cite{73} replaces the GRU with an LSTM to retain more high-frequency information. However, these methods~\cite{50,73} do not explain clearly why the frequency degradation phenomenon occurs during the iteration process.


We present the frequency convergence result of RAFT-stereo~\cite{52} by calculating the EPE evaluation metric for both high-frequency and low-frequency regions. Fig. \ref{fig:fengmian} (a) shows that the phenomenon of \textbf{frequency convergence inconsistency}, i.e., different frequency components exhibit different convergence behaviors. 
The underlying reason for the limited performance of iterative methods is that it jointly optimizes ``all frequency components'' using a single GRU update operator, without distinguishing between high and low frequencies. 
In this paper, \textbf{we introduce wavelets into stereo matching (Wavelet-Stereo) by explicitly decomposing the high-frequency and low-frequency components via discrete wavelet transform} as shown in Fig. \ref{fig:fengmian} (b), so that high and low-frequency features can be extracted and processed separately according to different frequency properties.

\begin{figure*}[t]  
\centering
\includegraphics[width=\textwidth]{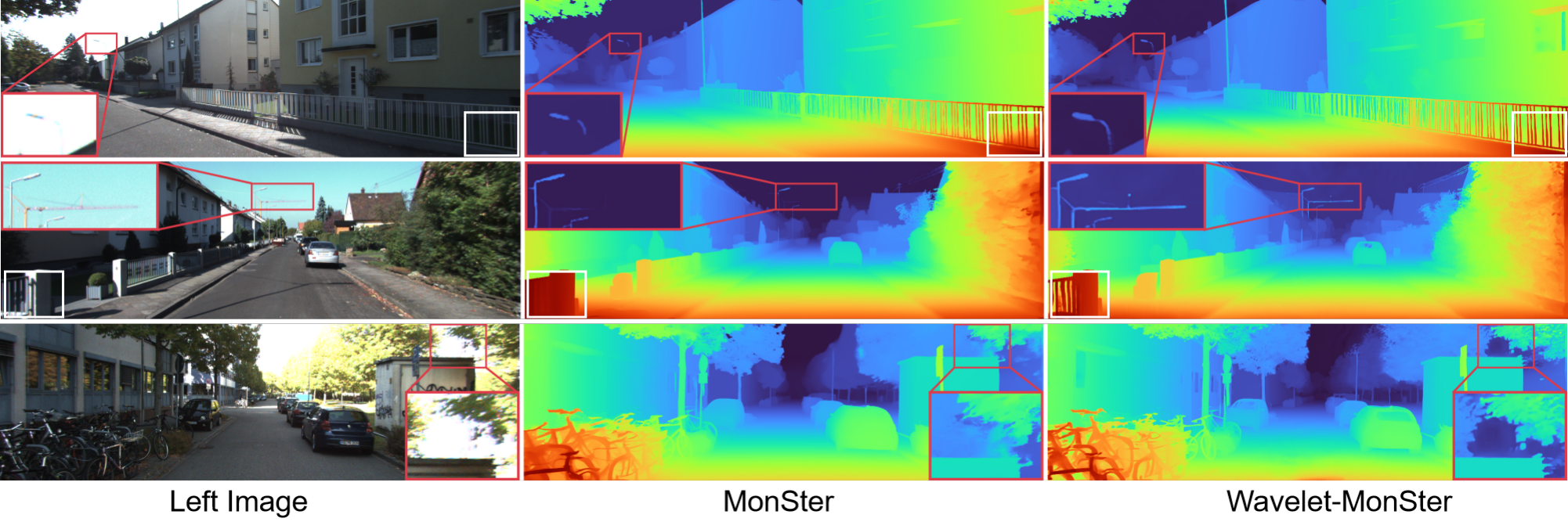}
\caption{\textbf{Visual comparison on KITTI.} All models are trained on Scene Flow and tested directly on KITTI~\cite{28,29}. Wavelet-MonSter outperforms MonSter in challenging areas with high-frequency details, fine structures.}
\label{fig:zero-shot}
\end{figure*}

Specifically, we first explicitly decompose an image into high and low frequency components using the Haar wavelet~\cite{25}. Then, we use two frequency feature extractors to extract multi-scale high-frequency features and low-frequency features separately. Finally, we propose a novel \textbf{high-frequency preservation update operator (HPU)} to simultaneously refines high-frequency information in edges and low-frequency information in smooth regions. The proposed HPU contains two modules: (1) An iterative-based frequency adapter can provide adaptive refined high-frequency features at different iteration steps by fine-tuning the initial high-frequency features. (2) A high-frequency preservation LSTM ensures that high-frequency information is not lost and also provides detailed texture information to guide low-frequency updates.
Our novel components can be plug-and-played into multiple iterative-based methods. Extensive experiments demonstrate that our Wavelet-Stereo outperforms the state-of-the-art methods and ranks $\mathbf{1^{st}}$ \textbf{on both the KITTI 2015 and KITTI 2012 leaderboards} for almost all metrics. Our framework can handle challenging scenes with fine details and textures in the distance, as show in Fig.\ref{fig:zero-shot}.

\section{Related Work}
\label{section:2}
\textbf{Aggregation-based methods in Stereo Matching.}  Aggregation-based methods~\cite{32, 34, 33, 35, 36, 37, 38, 39, 40, 75} have shown significantly improvement in accuracy and robustness. GCNet~\cite{36} a 4D correlation volume by concatenating the left and right feature maps within a predefined disparity search range, followed by cost aggregation using 3D convolutions to generate the final matching results. 
To avoid the use of 3D convolution, AANet~\cite{40} introduces intra-scale and cross-scale cost aggregation to capture the edge and non-edge area. 
ACVNet~\cite{xu2022attention} propose the attention concatenation volume to eliminate noise in the cost volume and improve the performance in the ambiguous region.
While aggregation-based methods achieve competitive accuracy, their computational overhead remains prohibitive for being applied to high-resolution inputs.

\textbf{Iterative-based methods in Stereo Matching.} Iterative-based methods~\cite{72, 71, 70, 69} have demonstrated significant advantages over aggregation-based methods. 
RAFT-Stereo~\cite{52} introduces an all-pairs correlation volume pyramid and utilizes GRU-based update operators to perform iterative disparity updates. On this basis, IGEV-Stereo~\cite{54} addresses the issue that the initial correlation volume is excessively coarse by a lightweight cost aggregation network before iteration. CREStereo~\cite{55} proposes a adaptive group correlation layer, computes correlations within local search windows to reduce memory and computational overhead. 
These methods suffer from slow convergence due to their inability to effectively coordinate the refinement of high and low frequency region.

\textbf{Frequency-based methods in Stereo Matching.} Although frequency domain information~\cite{46, 47, 48, 49} has been widely applied in computer vision tasks, its utilization in the field of stereo matching remains relatively limited. 
Selective-Stereo~\cite{50} proposes a selective recurrent unit module to adaptively captures multi-frequency information.
This data-driven frequency separation methods lack interpretability in their learned decomposition patterns and demonstrate limited generalization capability across diverse datasets.
DLNR~\cite{73} decouples the hidden state from the update matrix of disparity map and transfers high-frequency across iteration. However, it just defines the hidden state as a high-frequency feature which lacks a clear physical definition.

\section{Methodology}
\label{section:3}
For iterative-based stereo matching frameworks, the correlation volume $C$ is provided as a condition to the update operator, which predicts a series of disparity fields $\left\{d_1, \dots, d_{n_k}\right\}$ from a pair of rectified images $(I_L,I_R \in \mathbb{R}^{H \times W \times 3})$ by using the current estimate of disparity $d_k$ to index the correlation volume $C$.
In fact, the correlation volume containing considerable noisy information is used to predict $d_k$, causing the hidden state to lose critical information during the iteration process. For example, the hidden state increasingly contain global low-frequency information while losing local high-frequency information such as edges and thin objects~\cite{50,54,73}. Therefore, we introduce wavelets into stereo matching (Wavelet-Stereo) and propose a novel high-frequency preservation update (HPU) operator containing two modules: (1) An iterative-based frequency adapter can provide adaptive refined high-frequency features at different iteration steps by fine-tuning the initial high-frequency features $F_h$ extracted by a high-frequency feature extractor $E_h$. (2) A high-frequency preservation LSTM ensures that high-frequency information is not lost and also provides updated high-frequency guidance for updating the hidden state. 

\vspace{10pt}
\begin{figure}[htbp]
    \centering
    \includegraphics[width=\linewidth]{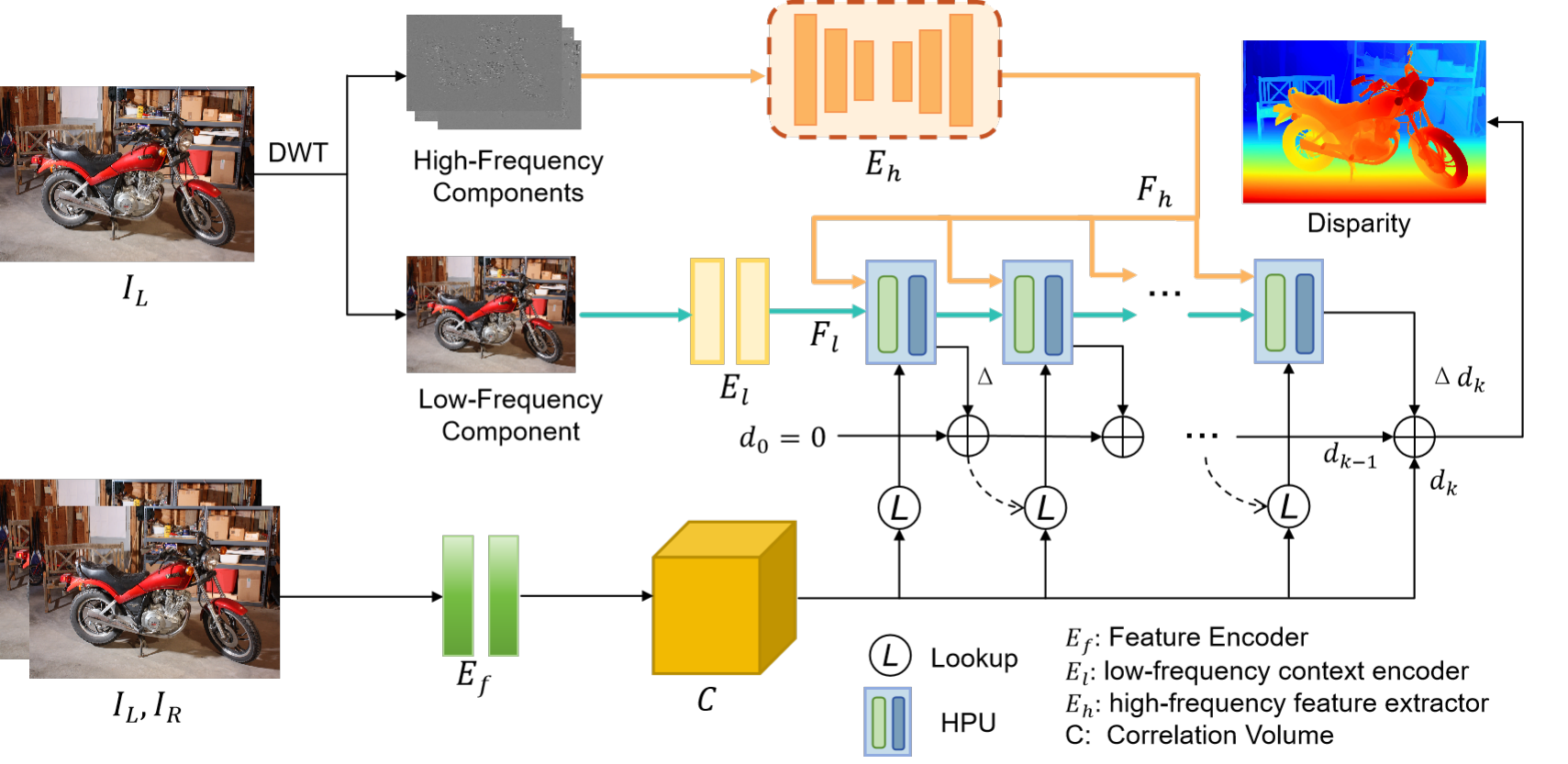}
    \caption{\textbf{Overview of Wavelet-RAFT}. Wavelet-RAFT employs a dual-branch architecture comprising: (1) a dedicated feature extraction branch for capturing high-frequency texture features $E_h$, (2) a update branch that progressively refines structural information. The aggregated high-frequency features $F_h$ serve as guidance information injected into the High-frequency Preservation Update (HPU) operator to update the hidden states during each iteration. }
    \label{fig:Overview} 
\end{figure}

\subsection{Overall Pipeline}
Since our method can be integrated into any iterative-based methods, we use Wavelet-RAFT as a representative example to demonstrate the key innovations of our framework, employing the same feature extraction network $E_\textit{f}$ and cost-volume construction as RAFT-Stereo~\cite{52} is used. As shown in Fig. \ref{fig:Overview}, our framework consists of three steps: (1) Frequency decomposition: We explicitly separate high-frequency and low-frequency components by discrete wavelet transform in Section \ref{subsec:DWT}. (2) Frequency Feature Extraction: we extract multi-scale high-frequency features and low-frequency features separately in Section \ref{subsec:MHFE}. (3) Iterative updating: we propose a novel update operator using high-frequency features as conditions to guide each iterative process in Section \ref{subsec:HPU}.

\subsection{Frequency Decomposition}
\label{subsec:DWT}

We use the Haar wavelet~\cite{25} to decompose the left image $I_L$ into four sub-images $I_{sub}$ with low and high frequency components, i.e., $I_{sub} = \mathrm{DWT}(I_{L})$, where $sub \in \{LL,LH,HL,HH\}$, $I_{LL}$ represents the low-frequency component, and $I_{LH}, I_{HL}, I_{HH}$ correspond to the high-frequency components. 
To obtain multi-scale frequency components, we iteratively apply DWT to the low-frequency sub-image ($I_{LL}$), i.e., $I_{sub}^{i}= DWT(I_{LL}^{i-1})$, where $i \in \{1,...,n_i\}$, $n_i=3$ is defined as the iteration number of DWT, $I_{sub}^i\in \mathbb{R}^{\frac{H}{2^i}  \times \frac{W}{2^i}  \times 3}$, and $I_{LL}^0=I_L$. 

\subsection{Multi-scale Frequency Feature Extraction}
Unlike Selective Stereo~\cite{50} relies on a channel-spatial attention module that implicitly learns a frequency-aware feature, we explicitly obtain the high and low frequency components of $I_L$ by DWT. This allows that high and low frequency features can be extracted and processed separately according to different frequency properties.

\begin{figure}[htbp]
    \centering
    \includegraphics[width=0.9\linewidth]{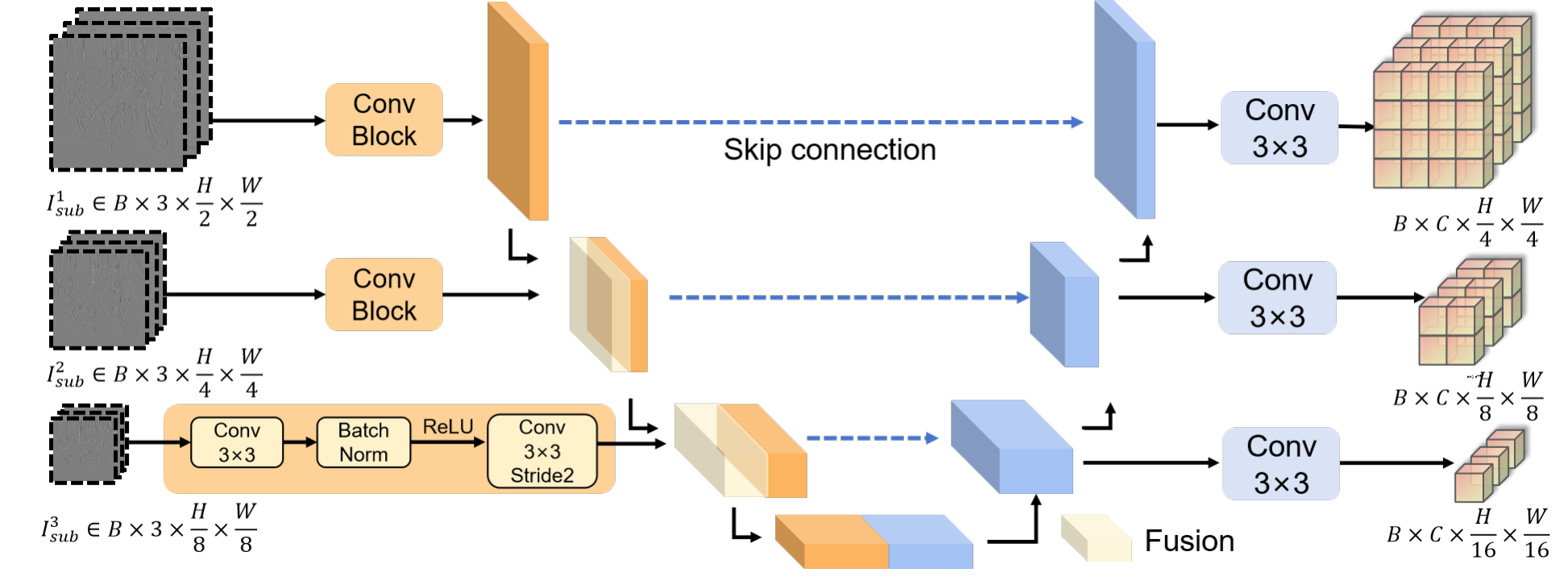}
    \caption{The franework of proposed high-frequency feature extractor consisting of a U-shaped network and a series of convolutions blocks, effectively capturing high-frequency feature through multi-scale feature aggregation and skip connection. }
    \label{fig:HAM}
\end{figure}

\textbf{High-frequency Feature Extraction.}\label{subsec:MHFE} To capture high-frequency details in textures, edges, and thin objects (see the second row of Fig. \ref{fig:zero-shot}), we design a U-shaped network as the high-frequency feature extractor $E_h$, as shown in Fig. \ref{fig:HAM}. It takes multi-scale high-frequency components $I_{sub}^i$ ($sub \in \{LH,HL,HH\}$) and outputs multi-scale high-frequency features $F_h^i$, i.e., $F_h^i=E_h(I_{sub}^i)$. The resolution of $F_h^i$ is halved by a convolutional block with a stride of 2, therefore $F_h^i$ contains $1/4$, $1/8$, $1/16$ scales.

\textbf{Low-frequency Feature Extraction.} To capture low-frequency information in smooth regions (see the third row of Fig. \ref{fig:zero-shot}), we use the context encoder in RAFT-Stereo~\cite{52} as the low-frequency feature extractor $E_l$. The network consists of a series of residual blocks and downsampling layers, producing multi-scale low features $F_l^i$ at 1/4, 1/8 and 1/16 resolution from low-frequency component $I_{LL}^1$, i.e., $F_l^i = E_l(I_{LL}^1)$.




\subsection{High-frequency Preservation Update Operator}
\label{subsec:HPU}

\begin{figure}[htp]
    \centering
    \includegraphics[width=0.95\linewidth]{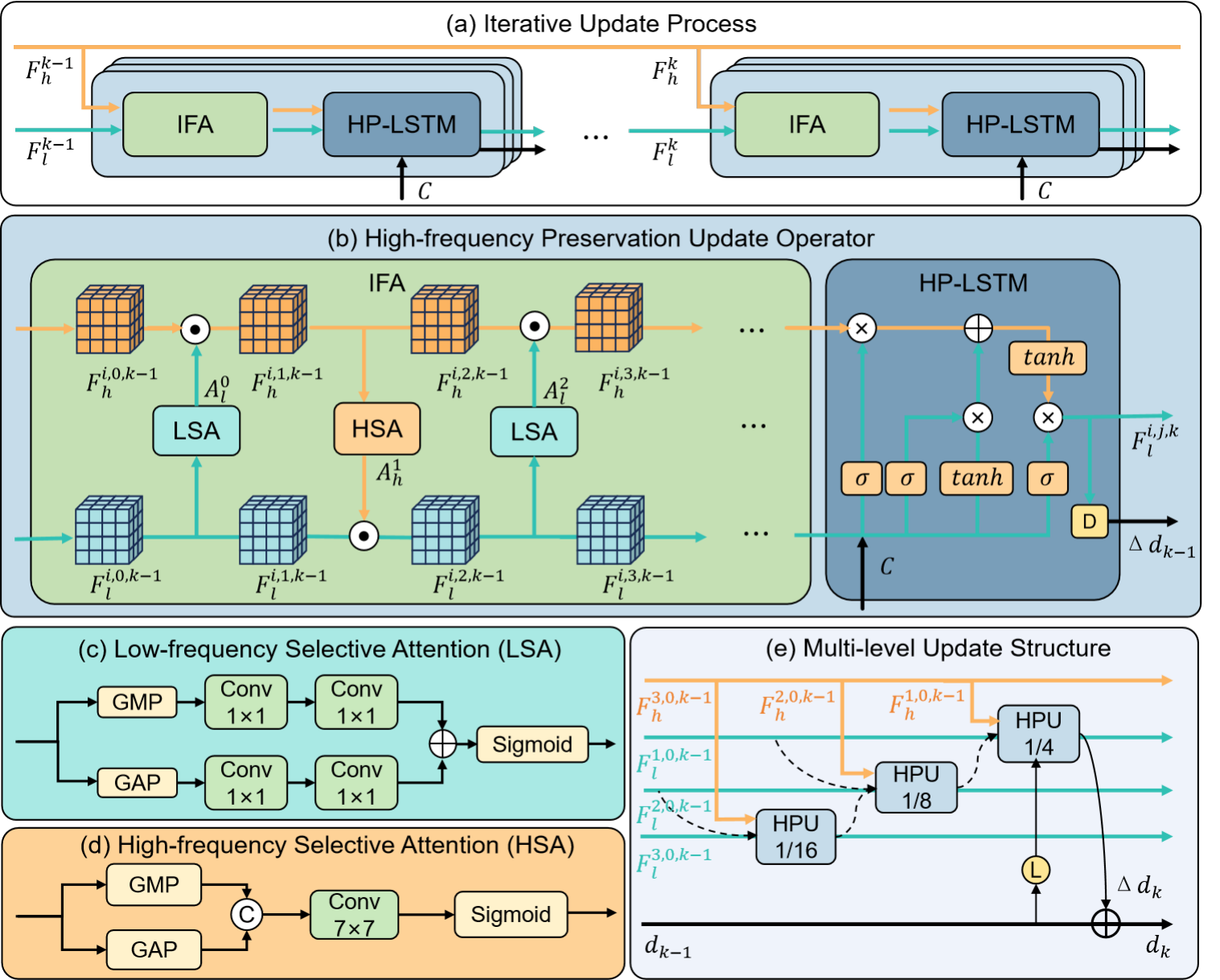}
    \caption{(a) The iterative update process of hidden states $F_l$, guiding by the aggregated high-frequency $F_h$. (b) Proposed high-frequency preservation update operator
    that finetunes the high-frequency in iterative-based frequency adapter and update hidden states by high-frequency preservation LSTM. (c) Low-frequency selection attention module adaptively integrates low-frequency contextual information to enhance high-frequency features (d) High-frequency selection attention module injects high-frequency attention maps to enrich low-frequency features. (e) Our multi-level update structure to update hidden states from 1/16 to 1/4.}
    \label{fig:HPU}
\end{figure}

Although the multi-scale high-frequency features have been sufficiently aggregated, directly incorporating them into the update operator is suboptimal, as the network focuses on different contents at different iteration stages.
To address this, we propose a novel high-frequency preservation update operator (HPU) containing an iterative-based frequency adapter (IFA) and a high-frequency preservation LSTM (HP-LSTM), as shown in Fig. \ref{fig:HPU}. Our IFA establishes a bidirectional interaction pathway that enables multiple rounds information fusion between the high-frequency features $F_h$ and the current hidden state $F_l$, achieving adaptive feature interaction through cross-complementarity.

\textbf{Iterative-based Frequency Adapter.} We design two attention modules to obtain adaptive refined frequency features at different iteration steps. (1) A low-frequency selection attention (LSA) module produces structural attention maps $A_l$ that provide global context to the high-frequency features $F_h$. (2) A high-frequency selection attention (HSA) module generates texture-aware attention maps $A_h$ to inject fine details into hidden states $F_l$. 
\begin{align}
F_h^{i,j,k}=A_l^{j-1}\odot F_h^{i,j-1,k}, \quad F_l^{i,j,k}=F_l^{i,j-1,k}, \quad  A_l^{j-1} = LSA(F_l^{i,j-1,k}) , j \in [1,3,5,...]\\
F_l^{i,j,k}=A_h^{j-1}\odot F_l^{i,j-1,k}, \quad F_h^{i,j,k}=F_h^{i,j-1,k}, \quad  A_h^{j-1} = HSA(F_h^{i,j-1,k}), j \in [2,4,6,...]
\label{Eq:5}
\end{align}
where $\odot $ represents elementwise multiplication, $i$ denotes the resolution dimension ($1/4$, $1/8$, and $1/16$), $j$ is the dimension in the IFA iteration, and $k$ is the dimension of the HPU iteration. $n_j$ is defined as the iteration number in IFA, while $n_k$ is defined the number of iterative updates.

\textbf{High-frequency Preservation LSTM:}
Existing iterative-based methods employ GRU-based update operators to update the full-frequency hidden states~\cite{52,50,54}, 
which results in degradation of high-frequency information during the iteration process. 
DLNR~\cite{73} introduces a LSTM to gradually decouple high-frequency information in updates, i.e., $F_l^k, F_h^k=LSTM(F_l^{k-1},F_h^{k-1})$. 
Since we have obtained the iteration-specific high-frequency features ($F_h^{i,j,k}\ne F_h^{i,j-1,k}$) after IFA,
we propose a high-frequency preservation LSTM, i.e., $F_h^{i,0,k}=F_h^{i,0,k-1}$.
and inject the high-frequency features into HP-LSTM as condition (others include the correlation volume $C$ and disparity $d_{k-1}$) for updating the current hidden state $F_l^k$.
\begin{align}
F_l^{k}, \triangle {d}_k &= {LSTM}_{HP}(F_l^{k-1} \vert  {F_{h}^{k-1}},L(C,{d_{k-1}}))
\end{align}

where $L$ refers lookup operator, the residual disparity $\triangle d_k$ is decoded from the hidden state $F_l^k$ by a decoder head $D$, i.e., $\triangle d_k=D(F_l^{k})$. The disparity $d$ is updated by
\begin{align}
{d}_{k} = {d}_{k-1} + \triangle {d}_k.
\end{align}

\subsection{Loss Function}
We use progressively weighted $L_1$ loss across all predicted disparities $\{d_k\}$. Given the ground truth of disparity $d_{gt}$, the total loss is defined as ($\gamma = 0.9$):
\begin{align}
\mathcal{L} = \sum_{k=1}^{n_k} \gamma^{n_k-i} \|{d}_k -  {d}_{gt}\|_1.
\end{align}

\section{Experiment}
\label{section:4}
\subsection{Implementation Details}
Wavelet-Stereo is implemented in Pytorch and trained using two NVIDIA A6000 GPUs. For all experiments, we use the AdamW~\cite{26} optimizer and clip gradients to the range [-1, 1]. We use the one-cycle learning rate schedule with a minimum learning rate of 2e-4. We train Wavelet-Stereo on the Scene Flow dataset~\cite{27} as the pretrained model with a batch size of 8 and 200k iterations. The ablation experiments are trained with a batch size of 6 for 100k steps.
We randomly crop images to 320 × 736 and use the same data augmentation as \cite{52} for training. We use 22 update iterations during training and 32 updates for evaluation.
The pipeline comparison of traditional iterative-based framework with ours is shown in Algorithm \ref{pipeline1}
and \ref{pipeline2}.

\begin{minipage}[t]{0.46\linewidth}
\begin{algorithm}[H] 
\caption{Traditional terative framework}
\begin{algorithmic}[1]
\label{pipeline1}
\REQUIRE a pair of rectified images $I_L, I_R$
\STATE $f_L,f_R=E_f(I_L,I_R)$
\STATE $C=\text{correlation}(f_L,f_R), d_0=0$
\STATE
\STATE $F_{l}^0=E_l(I_L)$
\STATE
\FOR{$k = 1, \cdots, n_k$}
    \STATE $F_l^{k}, \triangle d_k = GRU(F_l^{k-1}, \text{L}(C, d_k))$
    \STATE $d_{k} = d_{k-1} + \triangle d_k$
\ENDFOR
\RETURN disparity $d$
\end{algorithmic}
\end{algorithm}
\end{minipage}
\hfill
\begin{minipage}[t]{0.52\linewidth}
\begin{algorithm}[H] 
\caption{Ours}
\begin{algorithmic}[1]
\label{pipeline2}
\REQUIRE a pair of rectified images $I_L, I_R$
\STATE $f_L,f_R=E_f(I_L,I_R)$
\STATE $C=\text{correlation}(f_L,f_R), d_0=0$
\STATE $ \color{red}{I_{LL}^i,I_{HL}^i,I_{LH}^i,I_{HH}^i=DWT(I_L),i=1,2,3}$
\STATE $F_l^{0}=E_l(I_{LL}^1)$
\STATE $\color{red}{F_{h}=E_h(concat(I_{HL}^i,I_{LH}^i,I_{HH}^i))}$
\FOR{$k = 1, \cdots, n_k$}
    \STATE $\color{red}{F_l^{k},\triangle d_k = {HPU}(F_l^{k-1}, F_h, \text{L}(C, d_k))}$
    \STATE $d_{k} = d_{k-1} + \triangle d_k$
\ENDFOR
\RETURN disparity $d$
\end{algorithmic}
\end{algorithm}
\end{minipage}

\begin{figure*}[b]
\centering
\vspace{10pt}
\includegraphics[width=\textwidth]{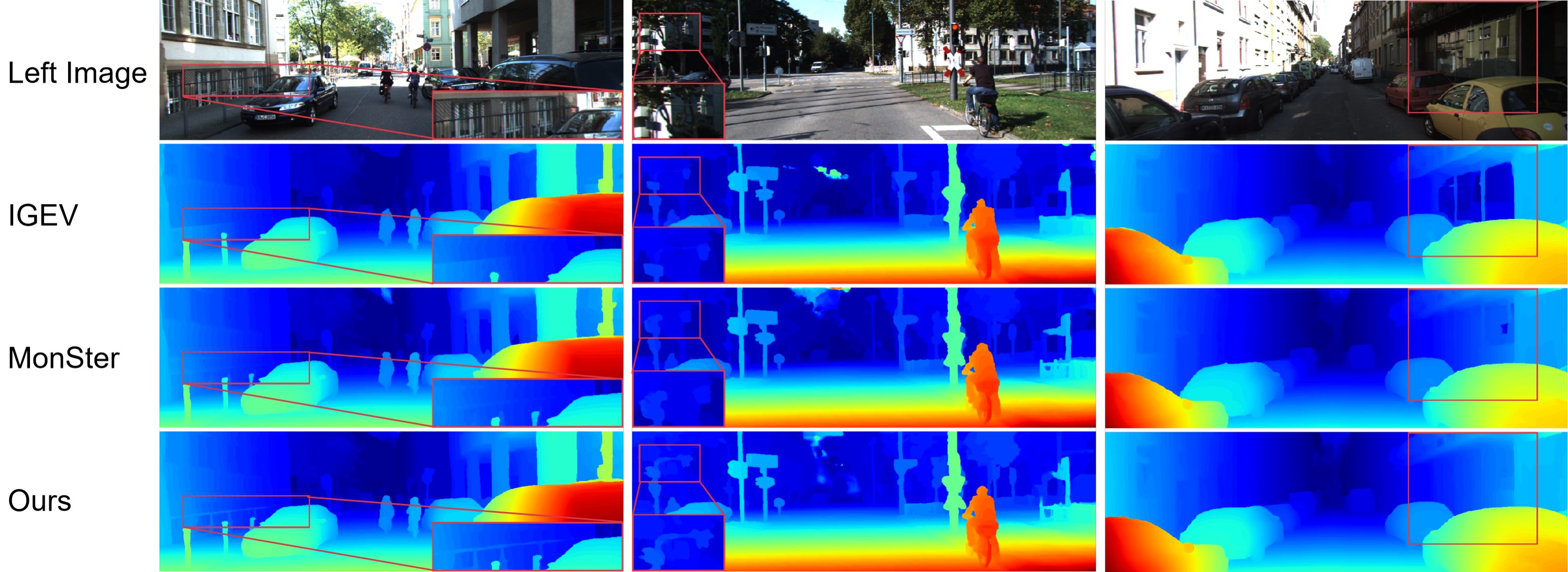}
\caption{Qualitative results on KITTI test set. Our Wavelet-MonSter outperforms MonSter in challenging areas with high-frequency details and weak texture.}
\label{fig:kitti_test}
\end{figure*}

\subsection{Benchmark datasets and Performance }

We evaluate Wavelet-Stereo on four widely used benchmarks and submit the results to online leaderboards for public comparison: KITTI 2012~\cite{28}, KITTI 2015~\cite{29}, ETH3D~\cite{30}, and Scene Flow~\cite{27}.

\textbf{Scene Flow~\cite{27}}. 
To verify the universality of our proposed framework, we take RAFT-Stereo and MonSter as baseline and integrate our framework. As shown in Table.\ref{tab:scene-flow}, both of our models surpass its baseline and our Wavelet-MonSter establishing a new state-of-the-art EPE benchmark on Scene Flow.

 To validate the ability of our method to handle different frequency regions, we split Scene Flow test set into high-frequency region and low-frequency region with Canny operator. As shown in Tab. \ref{tab:frequency}, 
  quantitative comparisons reveal that our Wavelet-Raft outperforms Selective-RAFT on EPE metric and surpasses the baseline by 22\%.
  Compared to Selective-IGEV~\cite{50} and DLNR~\cite{73}, which is also specifically designed to address different frequency regions, 
  our Wavelet-MonSter outperforms them  by 25.89\% and 10.3\% in high-frequency regions, 30.2\% and 13.7\% in low-frequency regions, respectively.

\textbf{ETH3D~\cite{30}}. Following CREStereo~\cite{55} and GMStereo [37], we firstly finetune the Scene Flow pretrained model on the mixed Tartan Air~\cite{wang2020tartanair}, CREStereo Dataset~\cite{55}, Scene Flow~\cite{27}, Sintel Stereo~\cite{butler2012naturalistic}, InStereo2k ~\cite{bao2020instereo2k} and ETH3D~\cite{30} datasets for 300k steps. Then we finetune it on the mixed CREStereo Dataset, InStereo2k and ETH3D datasets with for another 90k steps. As shown in Tab. \ref{online}, our Wavelet-MonSter outperforms MonSter by 4.4\% on Bad 1.0 metric, and achieves state-of-the-art performance among all published methods.

\begin{table}[t]
  \caption{Quantitative evaluation on Scene Flow test set. \textbf{Bold}: Best}
  \footnotesize
  \label{tab:scene-flow}
  \centering
  \scriptsize
  \setlength{\tabcolsep}{3.1pt} 
  \renewcommand{\arraystretch}{1.0} 
  \begin{tabular}{lccccccc}
    \toprule
    \textbf{Method}    & RAFT-Stereo~\cite{52} & ACVNet~\cite{65}  & IGEV-Stereo~\cite{54} & \textbf{Wavelet-RAFT (Ours)} & MonSter~\cite{91} & \textbf{Wavelet-MonSter (Ours)} \\
    \midrule
    \textbf{EPE (px)} & 0.53 & 0.48 & 0.47  & 0.46 & 0.37 & \textbf{0.36} \\
    \bottomrule
  \end{tabular}
\end{table}
\begin{table*}[tp]
\centering
\caption{Results on three popular benchmarks. All results are derived from official leaderboard publications or corresponding papers. All metrics are presented in percentages, except for RMSE, which is reported in pixels. For testing masks, ``All'' denotes testing with all pixels while ``Noc'' denotes testing with a non-occlusion mask. The \colorbox{red}{best} and \colorbox{yellow}{second best} are marked with colors.}
\resizebox{\textwidth}{!}{%
\begin{tabular}{l|ccc|cccc|cccc}
\toprule
& \multicolumn{3}{c|}{ETH3D~\cite{30}} & \multicolumn{4}{c|}{KITTI 2015~\cite{29}} & \multicolumn{4}{c}{KITTI 2012~\cite{28}} \\

\cmidrule(lr){2-4} \cmidrule(lr){5-8} \cmidrule(lr){9-12}
         & Bad1.0 & Bad1.0 & RMSE & D1-fg & D1-all & D1-fg & D1-all & Out-2 & Out-2&Out-3  & Out-3\\
         & Noc & All & Noc & Noc & Noc & All & All & Noc & All & Noc  & All\\
\hline
GwcNet~\cite{35}(CVPR 2019) & 6.42 & 6.95 & 0.69 & 3.49 & 1.92 & 3.93 & 2.11 & 2.16 & 2.71 & 1.32 & 1.70 \\
GANet~\cite{75}(CVPR 2019) & 6.22 & 6.86 & 0.75 & 3.37 & 1.73 & 3.82  & 1.93 & 1.89 & 2.50 & 1.19 & 1.60\\
LEAStereo~\cite{89}(NeurIPS 2020) & - & - & - & 2.65 & 1.51 & 2.91  &1.65 & 1.90 & 2.39 & 1.13& 1.45 \\
ACVNet~\cite{65}(CVPR 2022) & 2.58 & 2.86 & 0.45 & 2.84 & 1.52 & 3.07 & 1.65& 1.83 & 2.35 & 1.13 & 1.47 \\
CREStereo~\cite{55}(CVPR 2022) & 0.98 & 1.09 & 0.28 & 2.60 & 1.54 & 2.86 & 1.69 & 1.72 & 2.18 & 1.14 & 1.46\\
IGEV~\cite{54}(CVPR 2023) & 1.12 & 1.51 & 0.34 & 2.62 & 1.49 & 2.67 & 1.59 & 1.71 & 2.17 & 1.12 & 1.44\\
CroCo-Stereo~\cite{67}(ICCV 2023) & 0.99 & 1.14 & 0.30 & 2.56 & 1.51 & 2.65  &1.59 &- & - & - & - \\
Selective-IGEV~\cite{50}(CVPR 2024
) & 1.23 & 1.56 & 0.29 & 2.55 & 1.44 & 2.61 & 1.55 & 1.59 & 2.05 & 1.07 & 1.38\\
LoS~\cite{96}(CVPR 2024) & 0.91 & 1.03 & 0.31 & 2.66 & 1.52 & 2.81 & 1.65 & 1.69 & 2.12& 1.10 & 1.38 \\
NMRF-Stereo~\cite{95}(CVPR 2024) & - & - & - & 2.90 & 1.46 & 3.07 & 1.57 & 1.59 & 2.07 & 1.01 & 1.35\\
DEFOM-Stereo~\cite{92}(CVPR 2025) & 0.70 & 0.78 & 0.22 & \colorbox{red}{2.24}  & 1.33 & \colorbox{red}{2.23} & 1.41 & 1.43  & 1.79 & 0.94 & 1.18 \\
MonSter~\cite{91}(CVPR 2025) & \colorbox{yellow}{0.46} & \colorbox{yellow}{0.72} & \colorbox{red}{0.20} &   {2.76} & \colorbox{yellow}{1.33} & {2.81} & \colorbox{yellow}{1.41}& \colorbox{yellow}{1.36} & \colorbox{yellow}{1.75} & \colorbox{yellow}{0.84} & \colorbox{yellow}{1.09}\\
Wavelet-MonSter(ours)  & \colorbox{red}{0.44} & \colorbox{red}{0.68} & \colorbox{red}{0.20} &     \colorbox{yellow}{2.60} & \colorbox{red}{1.31} & \colorbox{yellow}{2.60} & \colorbox{red}{1.38} & \colorbox{red}{1.32} & \colorbox{red}{1.71} & \colorbox{red}{0.83} & \colorbox{red}{1.07}\\
\bottomrule
\end{tabular}
\label{online}
}
\end{table*}

\textbf{KITTI~\cite{28,29}}. Following the training of MonSter~\cite{91}, we finetune our pretrained model on the mixed dataset of KITTI 2012~\cite{28} and KITTI 2015~\cite{29} with a batch size of 8 for 50k steps. For best performance, we evaluate our Wavelet-MonSter on the test set of KITTI 2012 and KITTI 2015, with results submitted to the official KITTI online leaderboard. As shown in Table.\ref{online}, our Wavelet-MonSter achieves the best performance among all published approaches to date and ranks \textbf{$1^{st}$} on both the KITTI 2015 and KITTI 2012 leaderboards for almost all metrics, outperforming over 280 competing methods. 
Fig. \ref{fig:kitti_test} shows qualitative results on KITTI 2012 and KITTI 2015 test sets, where our Wavelet-MonSter significantly outperforms MonSter in both detailed high-frequency regions (see the first and second row of figure) and non-textured reflective regions (see the third row of figure) in the difficult scenarios.

\begin{table}[b]
\centering
\caption{Quantitative evaluation on Scene Flow test set in different regions (EPE).}
\begin{tabular}{lcccc}
\toprule
Method & High-frequency region & {Low-frequency region} \\
\midrule
RAFT-Stereo~\cite{52} & 34.00 & 0.72  \\
Selective-RAFT~\cite{50} & 27.89 & 0.57 \\
\textbf{Wavelet-RAFT} & \textbf{26.48} & \textbf{0.56} \\
\hline
DLNR~\cite{73} & 31.60 & 0.63 \\
Selective-IGEV~\cite{50} & 26.10 & 0.51\\
MonSter~\cite{91} & 26.08 & 0.47\\
\textbf{Wavelet-MonSter} & \textbf{23.42} & \textbf{0.44} \\
\bottomrule
\end{tabular}
\label{tab:frequency}
\end{table}

\subsection{Ablation Study}
We conducted comprehensive ablation studies to validate the contribution of each component in our framework. Due to the simplified training settings, the quantitative results of ablation experiments differ from the comparison results described above.  We present the main results of ablation experiments, and more results can be found in Appendix A.



\textbf{Effectiveness of proposed modules.} The results in the Table.\ref{ablation} demonstrate that it is effective and necessary to propose frequency-specific module for features with distinct convergence characteristics.

To assess the importance of the high-frequency feature extractor $F_h$ , we replace $F_h$  with a simple two-layer convolutional network. Quantitative results (EPE increases from 0.52 to 0.56) demonstrate that a powerful feature extraction network is needed to adequately fuse high-frequency information at multiple scales.

\begin{table*}[t]
\centering
\caption{Ablation study of the effectiveness of proposed modules on Scene Flow test set. HPU denotes High-frequency Preservation Update operator, $F_h$ denotes high-frequency feature extractor.}
\renewcommand{\arraystretch}{1.2}
\resizebox{0.85\textwidth}{!}{%
\begin{tabular}{l|ccc|ccc}
\hline
{Model} & {GRU} & {HPU} & {$F_h$} & {EPE (px)} & {D1 (\%)} 
\\
\hline
Baseline (RAFT-Stereo) & \checkmark & & & 0.62 & 8.40  \\
\hline
w/o HPU & \checkmark &  & \checkmark & 0.58 & 7.29 \\
w/o $F_h$ & & \checkmark & & 0.56 & 6.72  \\
{Full model (Wavelet-RAFT)} & & \checkmark  & \checkmark & \textbf{0.52} & \textbf{6.21} \\
\hline
\end{tabular}%
}
\label{ablation}
\end{table*}

To verify the effectiveness of our proposed HPU operator, we replace it with a standard GRU module in RAFT-Stereo which simply concatenates high-frequency and low-frequency features. This modification results in performance degradation across all metrics (EPE increases from 0.52 to 0.58 and D1 increases from 6.21 to 7.29), due to GRU’s all-frequency uniform processing of different frequency features. It demonstrates that our High-frequency Preservation LSTM can flexibly utilize high-frequency information as additional conditional inputs to better balance the retention and updating of high-frequency features.

\textbf{Number of IFA iteration.} To determine the most appropriate interaction iteration in IFA, we conduct a systematic investigation of IFA interaction rounds by varying j from 1 to 6. As quantified in Table.\ref{tab:ifa_rounds}, performance exhibits a clear peak at r=4 iterations, with both under-interaction (j$<$4) and over-interaction (j$>$4) leading to degraded results. This suggests: (1) sufficient rounds are necessary for 
finetuning the iteration-specific high-frequency features,  yet (2) excessive iterations may cause feature over-smoothing.

\textbf{Number of Iterations.} The results in the Table.\ref{tab:efficiency} demonstrate that our framework can significantly accelerate the speed of network convergence, which is largely attributed to the fact that we address the convergence characteristics of different frequency features separately. Specifically, providing accurate high-frequency priors in the initial iteration allows the network to focus attention on global context optimization, achieving superior performance and significantly fewer iterations than traditional iterative-based methods.  
Our Wavelet-RAFT need only 8 iterations to surpass the performance of RAFT-Stereo  while reducing runtime by 38.6\%.
\begin{table}[ht]
    \centering
    \begin{minipage}[t]{0.4\linewidth}
        \centering
        \caption{Ablation study of the rounds(j) in IFA.}
        \label{tab:ifa_rounds}
        \begin{tabular}{ccc}
            \toprule
            {Rounds (j)} & {EPE} & {Runtime(s)}  \\
            \midrule
            1 & 0.394 &  0.680  \\ 
            2 & 0.383 &  0.686  \\
            3 & 0.371 &  0.772  \\
            4 & \textbf{0.367} & 0.790  \\
            5 & 0.371 &  0.865s \\
            6 & 0.373 &  0.875s\\
            \bottomrule
        \end{tabular}
    \end{minipage}%
    \hfill
    \begin{minipage}[t]{0.55\linewidth}
        \centering
        \caption{Ablation study of the number of iterations.}
        \label{tab:efficiency}
        \begin{tabular}{lcccc}
            \toprule
            Model  & Iteration  & EPE & Runtime (s) \\
            \midrule
             & 32 & 0.53 & 0.44 \\
            RAFT-Stereo~\cite{38}& 16 & 0.63 & 0.26\\
            & 12 & 0.53 & 0.22 \\
            \hline
              & 32 & \textbf{0.46} & 0.79 \\
            Wavelet-RAFT & 16 & 0.47 & 0.45 \\
               & 12 & 0.50 & 0.36 \\
               & 8 & 0.52 & 0.27 \\
            \bottomrule
        \end{tabular}
    \end{minipage}
\end{table}

\section{Conclusion}
\label{section:5}

We find that the underlying reason for the high-frequency degradation during the iteration process is the inconsistent convergence of the performance in the low-frequency and high-frequency regions. We propose a wavelet-based stereo matching framework (Wavelet-Stereo) for solving frequency convergence inconsistency. The core contribution is the proposed high-frequency preservation update operator, consisting of an iterative-based frequency adapter and a high-frequency preservation LSTM, which can simultaneously refine the high-frequency information at the edges and the low-frequency information in the smoothing region without losing the high-frequency information. Our novel components can be plug-and-played into multiple iterative-based methods. By processing high and low frequency components separately, our framework can handle challenging scenes with fine details and textures in the distance.
In the future, we will release a real-time version for better real-world deployment.

\bibliographystyle{unsrt}

\newpage

\section*{Appendix}

\subsection*{A \quad Dataset and evaluation metrics}
\textbf{Pretrain dataset:} Scene Flow~\cite{27} is a synthetic stereo matching dataset consisting of 35,454 training image pairs and 4,370 testing image pairs, with a resolution of 960×540. It provides dense disparity maps as ground truth annotations for each image pair. All models in this work are trained exclusively on the SceneFlow training dataset.

\textbf{Zero-shot and finetune datasets:} To validate the generalization capability of our model, we evaluate its performance on the training sets of the following four real-world datasets. \textbf{KITTI 2012}~\cite{28} and \textbf{KITTI 2015}~\cite{29} are real-world driving scene datasets. Specifically, KITTI 2012 provides 194 training pairs and 195 testing pairs, while KITTI 2015 offers 200 training pairs and 200 testing pairs. 
\textbf{ETH3D}~\cite{30} consists of gray-scale stereo pairs acquired from diverse indoor and outdoor scenes, comprising 27 pairs for training and 20 pairs for testing. \textbf{Middlebury}~\cite{31} provides 15 training pairs and 15 testing pairs of high-resolution stereo images captured in indoor environments. 

\textbf{Metrics:} As usual, we use end-point-error (EPE) and kpx for Scene Flow datasets evaluation metrics, where EPE is the average $l_{1}$ distance between the prediction and ground truth disparity. And kpx denotes the percentage of outliers with an absolute error greater than k pixels. Referencing previous studies, the thresholds set for each dataset are as follows: 3 pixels for KITTI-2012 and KITTI-2015, 2 pixels for Middlebury, and 1 pixel for ETH3D.


\subsection*{B \quad Implementation}
\subsubsection*{B.1 \quad Implementation Details}
Following~\cite{52}, all models are trained with the Adam optimizer ($\beta_1=0.9,\beta_2=0.999$). For data augmentation setting, the image saturation was adjusted between 0  and 1.4, the right image was perturbed to simulate imperfect rectification that is common in datasets such as ETH3D and Middlebury.
We froze all the batch normalization layers in training process. The maximum disparity $D$ for training and evaluation is set to $D = 192$.

\subsubsection*{B.2 \quad Frequency Convergence Inconsistency Experiment}
To quantitatively evaluate frequency-specific performance, we generate edge masks using the Canny operator (implemented via OpenCV, lower=100, upper=200) on ground truth (GT) images for explicit separation of  high-frequency regions and low-frequency regions. The binary edge map M serves as a high-frequency region mask, enabling calculation of high-frequency endpoint error (EPE) through element-wise multiplication:
\begin{equation}
    EPE_{high}=M \odot EPE_{total} 
\end{equation}

Conversely, $(1 - M)$ serves as a low-frequency region mask and the low-frequency error is computed using the inverted mask $(1 - M)$：

\begin{equation}
EPE_{low}=(1-M)\odot EPE_{total}
\end{equation}

\subsubsection*{B.3 \quad Structure of High-frequency Preservation Update Operator}
The High-frequency Preservation Update Operator is consisted of Iterative-based Frequency Adapter and High-frequency Preservation LSTM.

For Iterative-based Frequency Adapter, it contains two frequency attention module:  A low-frequency selection attention (LSA) module and
a high-frequency selection attention (HSA) module.
The LSA module processes low-frequency features carrying global structural information through a dual-path architecture. Let $F_l \in R^{H \times W \times C}$ denote the input low-frequency feature map. The module first applies both Global Max Pooling (GMP) and Global Average Pooling (GAP) along spatial dimensions to obtain channel-wise features. These pooled features then undergo channel transformation via 1×1 convolutions ($W_1, W_2 \in R^{C \times C}$) followed by $ReLU$ activation function：
\begin{equation}
\begin{aligned}
z_{max} &= ReLU[W_1(GMP(F_l))]\\
z_{avg} &= ReLU[W_2(GAP(F_l))]\\
A_L &= \sigma(z_{max} + z_{avg})
\end{aligned}
\end{equation}
where $\sigma$ denotes the sigmoid activation function. 

The HSA module targets high-frequency patterns containing local textures and details. It employs identical pooling operations but processes them through a 7×7 convolutional layer $W_3$ to capture broader spatial contexts while suppressing noise:
\begin{equation}
    A_H = \sigma(W_3(Concat(z_{max}, z_{avg})))
\end{equation}
where $\sigma$ denotes the sigmoid activation function and Concat denotes concatenating along the channel dimension. The LSA module provides global structural context to guide high-frequency processing, while the HSA module supplies local texture details to enrich low-frequency representations. 

For the High-frequency Preservation LSTM, it takes high-frequency feature $F_h$ as condition priors along with cost volume C, disparity $d_k$ to update the hidden states $F_l$：
\begin{equation}
\begin{aligned}
x_k &= [\text{Encoder}_g(\text{C}), \text{Encoder}_d(d_k), d_k]\\
i_t &= \sigma(\text{Conv}([h_{k-1}, x_k], W_i) + b_{hi})\\
f_t &= \sigma(\text{Conv}([h_{k-1}, x_k], W_f) + b_{hf})\\
g_t &= \tanh(\text{Conv}([h_{k-1}, x_k], W_g) + b_{hg})\\
o_t &= \sigma(\text{Conv}([h_{k-1}, x_k], W_o) + b_{ho})\\
c_t &= f_t * F_{h} + i_t * g_t\\
F_l &= o_t * \tanh(c_t)
\end{aligned}
\end{equation}

\subsection*{C \quad More Quantitative Results}

\begin{figure}[h]
    \centering
    \includegraphics[width=\linewidth]{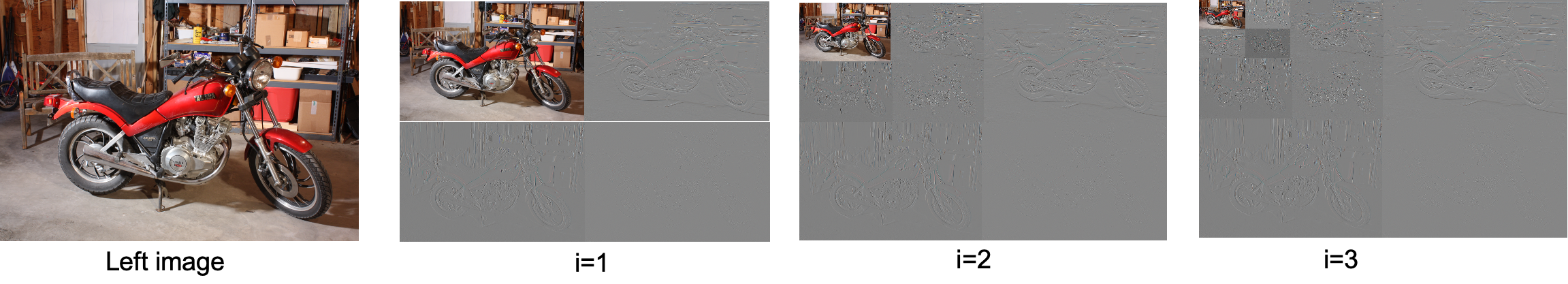}
    \caption{Different level of DWT decomposition (i refers DWT level). }
    \label{fig:dwt}
\end{figure}

\textbf{Effectiveness of multi-scale high-frequency feature extractor} To evaluate the efficacy of our multi-scale high-frequency feature extractor $F_h$, we conduct comprehensive ablation studies by feeding multi-level DWT outputs (Fig.\ref{fig:dwt}) into the module. 
It introduces only a minimal parameter increase through an efficient fusion module that aggregates multi-level high-frequency features from DWT decomposition.
Quantitative evaluation on the Scene Flow test set (Table. \ref{hfe}) demonstrates that this lightweight design adds just 0.77M additional parameters while achieving 2.3\% improvements (EPE decreases from 0.483 to 0.472).

\begin{table}[H]
\centering
\caption{Ablation studies of the effectiveness of our multi-scale high-frequency feature extractor. 1, 2, 3 stand for the level of Discrete Wavelet Transform (DWT).}
\begin{tabular}{lccc}
\toprule
\textbf{Method} & \textbf{EPE} & \textbf{D1} & \textbf{Params. (M)} \\
\midrule
HPU & 0.563 & 6.92 & 0.55 \\
HPU + HAM$_1$ & 0.483 & 6.39 & 4.36 \\
HPU + HAM$_2$ & 0.472 & 6.26 & 4.73 \\
HPU + HAM$_3$ (Ours) & \textbf{0.467} & \textbf{6.21} & \textbf{5.5} \\
\bottomrule
\end{tabular}
\label{hfe}
\end{table}

Our high-frequency feature extractor which is fed 3-level DWT decomposition outputs achieves effective fusion and utilization of multi-scale high-frequency features.
 This carefully balanced design maintains the model's compactness and practical deployability while enabling effective multi-scale high-frequency feature utilization.

\textbf{Parameter and Computational Analysis}
We further provide quantitative results on memory usage and computational cost. We use a single
Nvidia A6000 graphics card (with 48 GiB memory) and the batch size is set to 1 for the inference (16 iterations).
The memory consumption and
computational cost is shown in Table.\ref{tab:computational_complexity}
\begin{table}[t]
\centering
\caption{Computational complexity breakdown per stage. Runtime, GPU memory usage, number of parameters, and equivalent FPS are reported.}
\begin{tabular}{@{}lcccc@{}}
\toprule
Stage & Memory(MB) & Params(M) & Runtime(ms) \\ \midrule
DWT & 0 & - & 33.31  \\
Low-frequency Feature Extraction & 1660 & 4.32 & 10.65 \\
High-frequency Feature Extraction & 2064 & 7.01 & 5.44  \\
Cost volume & 2072 & - & 70.54  \\
HPU-Refinement & 2178 & 6.47 & 369.16  \\ \bottomrule
\end{tabular}
\label{tab:computational_complexity}
\end{table}

\subsection*{D \quad More Qualitative Results}

\begin{figure}[h]
    \centering
    \includegraphics[width=\linewidth]{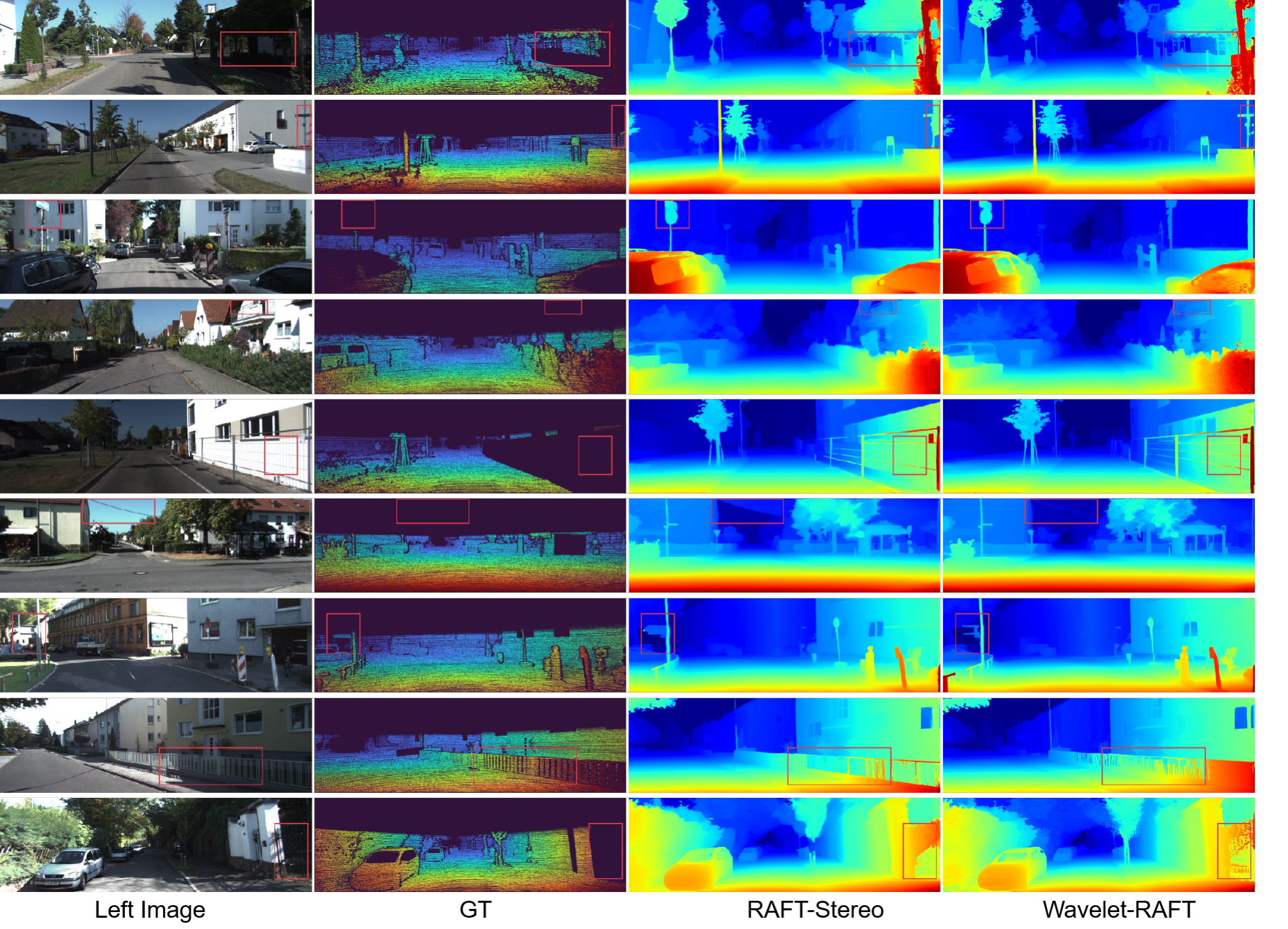}
    \caption{\textbf{Qualitative Results – Zero-Shot Generalization on the KITTI 2012 and  KITTI 2015 train sets}. }
    \label{fig:zero-kitti}
\end{figure}

\begin{figure}[h]
    \vspace{-10pt}
    \centering
    \includegraphics[width=0.95\linewidth]{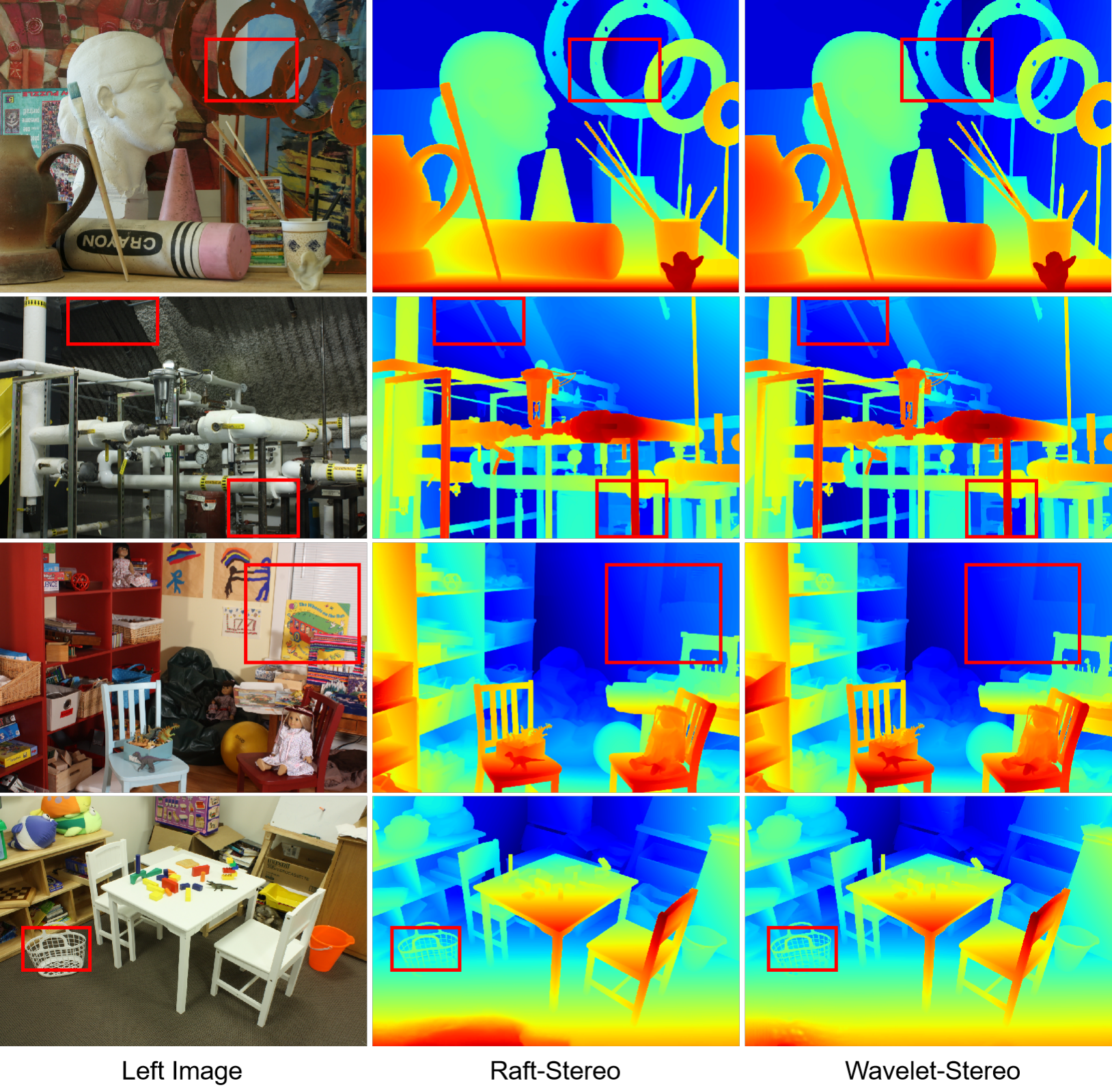}
    \caption{Qualitative Results – Zero-Shot Generalization on the Middlebury~\cite{31} train set.}
    \label{fig:zero-mb}
\end{figure}

\begin{figure}[H]
    \vspace{-10pt}
    \centering
    \includegraphics[width=0.95\linewidth]{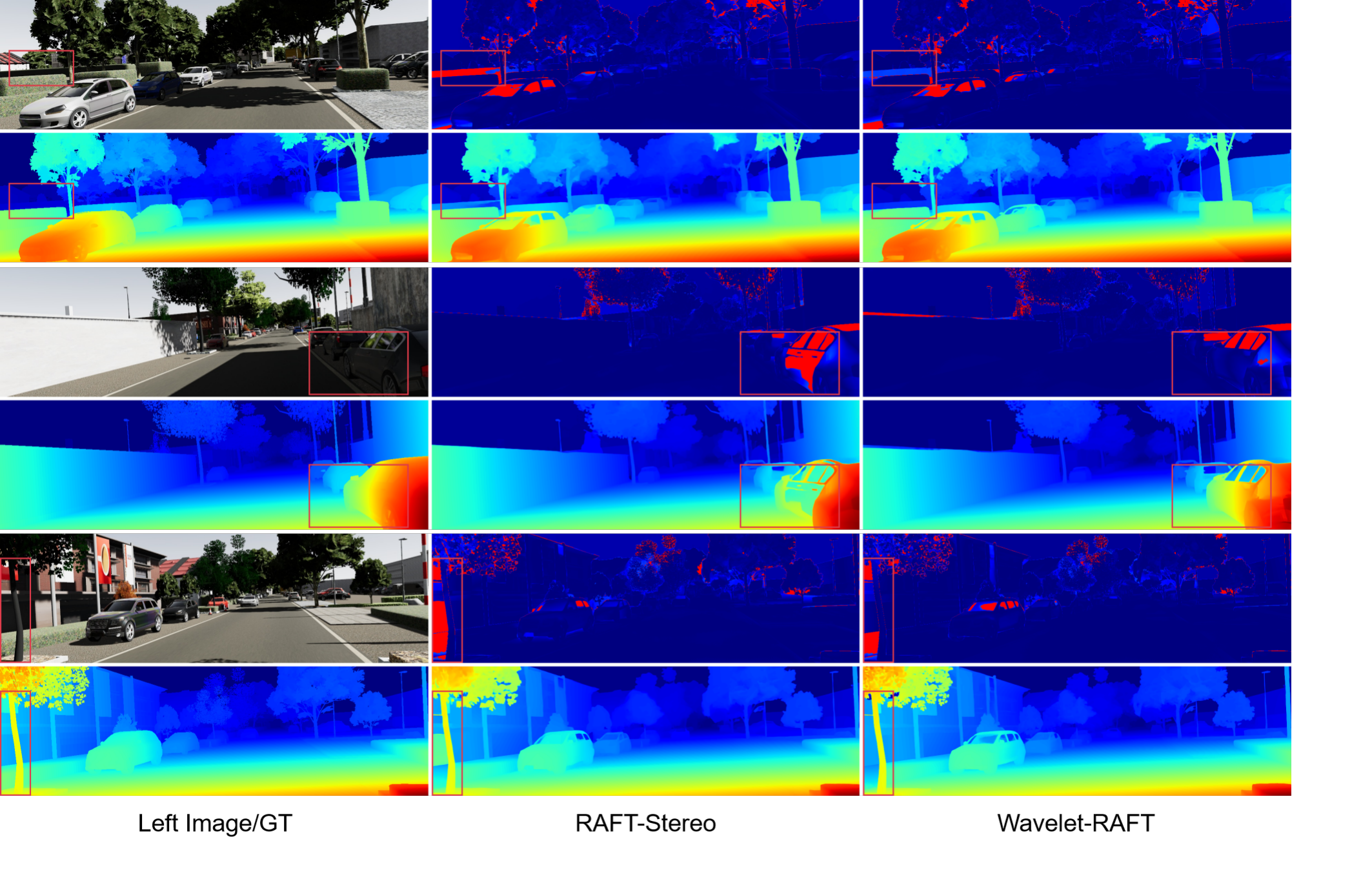}
    \caption{Qualitative results on VKITTi train set. The first column shows the left image and the corresponding ground-truth disparity map.
The rest columns show the error map and the predicted disparity map of RAFT-Stereo and Wavelet-RAFT, respectively. }
    \label{fig:vkitti}
\end{figure}

\begin{figure}[H]
    \centering
    \includegraphics[width=\linewidth]{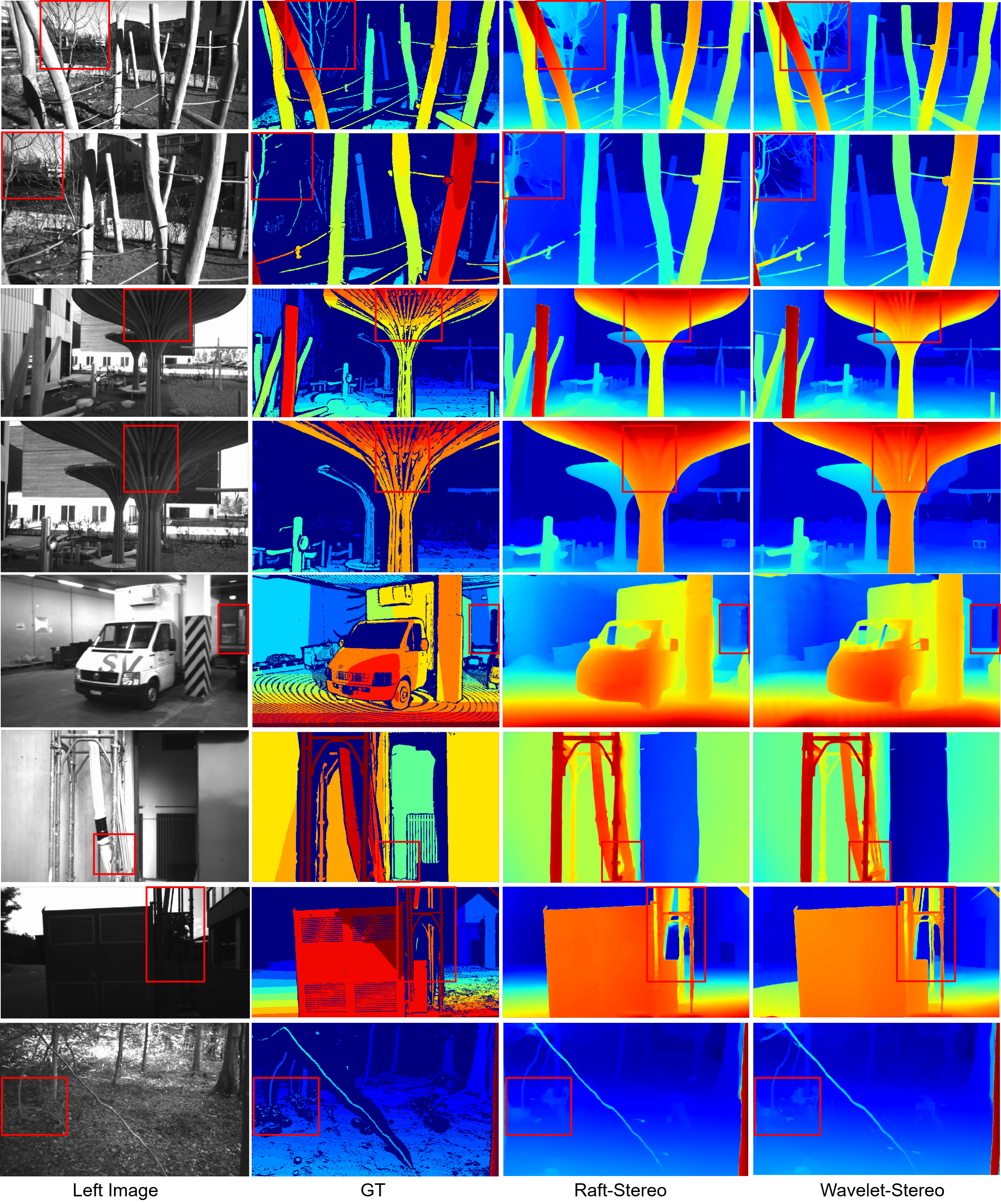}
    \caption{Qualitative Results – Zero-Shot Generalization on the ETH3D~\cite{30} train set}.
    \label{fig:zero-eth3d}
\end{figure}

\begin{figure}[h]
    \centering
    \includegraphics[width=\linewidth]{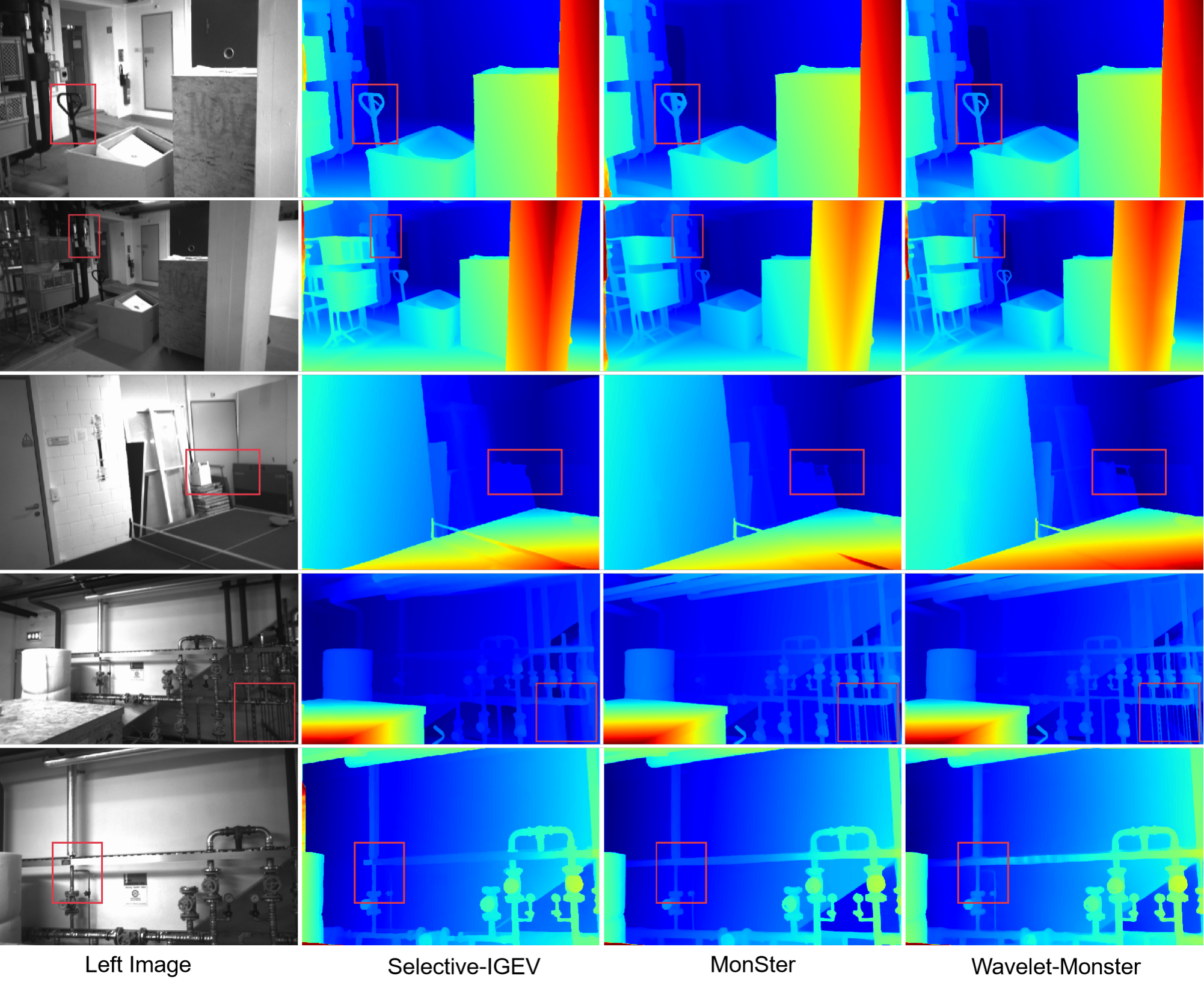}
    \caption{{Qualitative results on ETH3D test set.} }
    \label{fig:eth3d-finetune}
\end{figure}

In this section, we provide a comprehensive qualitative comparison between our method and the baselines on four widely used real-world datasets (KITTI 2012~\cite{28}, KITTI 2015~\cite{29}, Middlebury~\cite{31} and ETH3D~\cite{30}). As shown in Fig.\ref{fig:vkitti}, Fig.\ref{fig:zero-kitti}, Fig.\ref{fig:zero-eth3d} and Fig.\ref{fig:zero-mb}, our Wavelet-RAFT exhibits significantly superior zero-shot generalization performance compared to baseline model when pretrained exclusively on the synthetic SceneFlow~\cite{27} dataset. 
 Our Wavelet-MonSter demonstrates remarkable performance in preserving hierarchical details in the predicted disparity maps, with even the most delicate  structures being accurately maintained, as shown in Figure. \ref{fig:eth3d-finetune}.

\newpage
\begin{figure}[h]
    \centering
    \includegraphics[width=\linewidth]{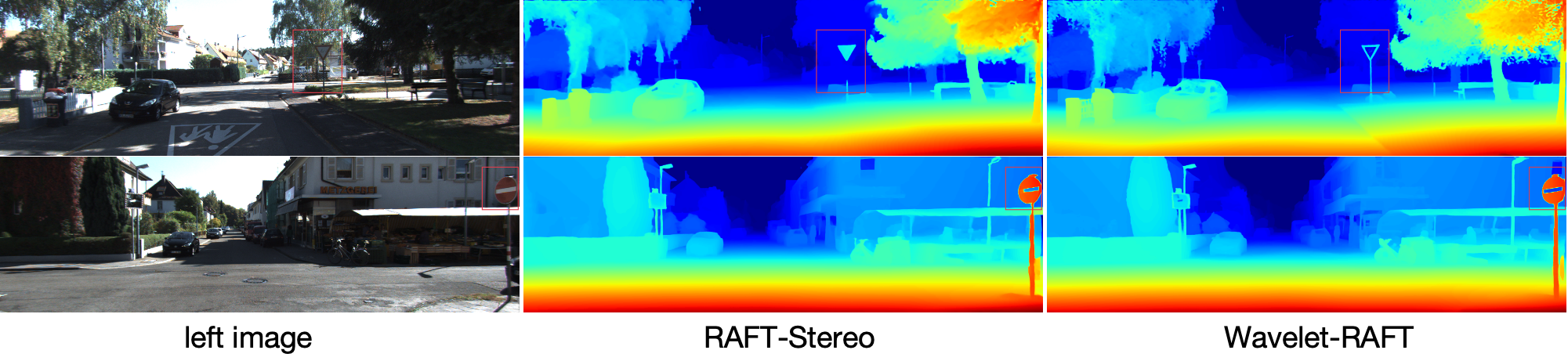}
    \caption{ Examples of failure cases for the proposed method. Poor performance due to unnecessary extraction of task-irrelevant information.}
    \label{fig:badcase}
\end{figure}

\subsection*{E \quad Discussions, Limitations, and Further Work}
\textbf{Limitations.} While the proposed method demonstrates strong performance, the computational overhead induced by the DWT decomposition, multi-scale feature extraction, and iterative frequency adapter (IFA) operations could potentially hinder real-time deployment. These limitations highlight important trade-offs between frequency-aware precision and computational practicality that warrant further investigation in future work. 

\textbf{Further Work.} Here are some directions of our future work.

1. Semantics-guided high-frequency processing pipeline that discriminatively extracts task-relevant high-frequency information.

2. Adaptive number of iteration  for different scenarios.

3. Exploring the application of diffusion model in stereo matching.
\end{document}